\documentclass[journal]{IEEEtran}
\usepackage[hidelinks]{hyperref}
\usepackage{cite}
\usepackage{amsmath,amssymb,amsfonts}
\usepackage{algorithmic}
\usepackage{graphicx}
\usepackage{textcomp}
\usepackage{xcolor}
\usepackage{colortbl}
\usepackage{amsfonts}
\usepackage{booktabs}
\usepackage{siunitx}
\usepackage{bbm}
\usepackage[T1]{fontenc}
\DeclareMathAlphabet{\mathcal}{OMS}{cmsy}{m}{n}
\SetMathAlphabet{\mathcal}{bold}{OMS}{cmsy}{b}{n}

\begin{document}
\title{Physics-Enhanced Graph Neural Networks For Soft Sensing in Industrial Internet of Things}
\author{Keivan Faghih Niresi,
        Hugo Bissig,
        Henri Baumann,
        and~Olga Fink,~\IEEEmembership{Member,~IEEE}
\thanks{(\emph{Corresponding author: Olga Fink.})}
\thanks{Keivan Faghih Niresi and Olga Fink are with Intelligent Maintenance and Operations Systems (IMOS) Lab, EPFL, Switzerland,
e-mail: (keivan.faghihniresi@epfl.ch, olga.fink@epfl.ch)}
\thanks{Hugo Bissig and Henri Baumann are with Swiss Federal Institute of Metrology (METAS).}
}

\maketitle

\begin{abstract}
The Industrial Internet of Things (IIoT) is reshaping manufacturing, industrial processes, and infrastructure management. By fostering new levels of automation, efficiency, and predictive maintenance, IIoT is transforming traditional industries into intelligent, seamlessly interconnected ecosystems. However, achieving highly reliable IIoT can be hindered by factors such as the cost of installing large numbers of sensors, limitations in retrofitting existing systems with sensors, or harsh environmental conditions that may make sensor installation impractical. Soft (virtual) sensing leverages mathematical models to estimate variables from physical sensor data, offering a solution to these challenges. Data-driven and physics-based modeling are the two main methodologies widely used for soft sensing. The choice between these strategies depends on the complexity of the underlying system, with the data-driven approach often being preferred when the physics-based inference models are intricate and present challenges for state estimation. However, conventional deep learning models are typically hindered by their inability to explicitly represent the complex interactions among various sensors. To address  this limitation, we adopt  Graph Neural Networks (GNNs), renowned for their ability to effectively  capture the complex  relationships between sensor measurements. In this research, we propose physics-enhanced GNNs, which integrate principles of physics into graph-based methodologies. This is achieved by augmenting additional nodes in the input graph derived from the underlying characteristics of the physical processes. Our evaluation of the proposed methodology on the case study of district heating networks reveals significant improvements over purely data-driven GNNs, even in the presence of noise and parameter inaccuracies. Our code and data are
available under https://github.com/EPFL-IMOS/PEGNN\_SS.

\end{abstract}


\section{Introduction}
\label{sec:introduction}

The Industrial Internet of Things (IIoT) has been rapidly emerging as an important framework, using interconnected sensors, actuators, and related devices not only for automation, but also for interconnectivity between subparts of the network, thereby facilitating decision-making processes within the industrial ecosystem. IIoT goes beyond traditional data collection, focusing instead on streamlining operations and facilitating seamless communication between subcomponents of the network. This paradigm not only drives digital transformation but also facilitates automation and optimizes efficiency across diverse sectors such as water, manufacturing, energy, infrastructure management, and healthcare \cite{wu2021graph, babayigit2023industrial}.  However, The reliability of IIoT strongly depends on the quality of data and information gathered through sensor networks integrated into complex systems \cite{xing2020reliability}. Therefore, achieving highly reliable measurements in IIoT may encounter obstacles such as the cost of installing large numbers of sensors, limitations in retrofitting existing systems with sensors, sparse deployment of sensors, malfunctioning sensors, energy constraints associated with deploying dense sensor networks, and harsh environmental conditions that may make sensor installation impractical \cite{jiang2021soft}. As a result, instead of solely relying on expensive hardware solutions to address these obstacles, they can be effectively addressed through computational sensing techniques, thereby unlocking the full potential of intelligent systems facilitated by IIoT. For instance, recent studies present frameworks that optimize sensors' sampling rates, edge servers' preprocessing modes, edge–cloud resource allocation, and joint sensor activation with mobile charging scheduling. These approaches can significantly reduce system-wide energy consumption \cite{shi2022joint, chen2022learning}.  

Soft (virtual) sensing presents a promising method for augmenting the capabilities of physical sensors, improving data quality, and enhancing the efficacy of IIoT in monitoring complex system operations \cite{shang2014data, yan2016data, yuan2018deep}. Specifically, a soft sensor functions as a mathematical model that estimates desired variables at new target locations using data collected from physical sensors. Soft sensors prove to be valuable for inferring variables in places where physical sensors are unavailable \cite{jiang2021soft}. In soft sensing, two primary approaches have been employed: data-driven and physics-based modeling.

Traditionally, first principle approaches and methods like the Kalman filter and observers have been used to estimate model parameters \cite{chen2015soft}. Standard Kalman filters are commonly  used within first-principles-based approaches for model parameter estimation \cite{guo2014development}. However, these approaches typically presume linearity in both the system and observation model equations \cite{schimmack2018extended}, whereas many real-world industrial processes exhibit significant nonlinearity. Therefore, adopting  a nonlinear process assumption has become crucial to overcome   this limitation and expand   the application of soft sensors in actual industrial scenarios\cite{yang2019kpi}. In such cases, alternative methods such as the unscented Kalman filter (UKF) \cite{wan2000unscented}, the square-root UKF (SR-UKF) \cite{van2001square}, and the extended Kalman filter (EKF) \cite{ribeiro2004kalman}, become feasible choices. Nevertheless, developing physics-based soft sensors requires an in-depth understanding of the underlying processes and significant effort for model development.

When the system's underlying physical processes pose challenges for state estimation, data-driven approaches are typically preferred \cite{yuan2020deep}. The development of data-driven soft sensing approaches has involved various multivariate statistical and machine learning approaches, including principal component regression (PCR), partial least squares regression (PLS), and support vector machines (SVM) \cite{kim2021review}. Deep learning techniques are particularly noteworthy for their ability to autonomously learn features, eliminating the need for laborious feature engineering -- this is especially advantageous in scenarios where a thorough understanding of domain-specific features is absent  \cite{shang2019data}. As a result, deep learning has emerged as one of the most widely used methods for data-driven soft sensing \cite{sun2021survey}. 

In principle, soft sensor modeling can  also be considered as time series missing data imputation problem, a domain in which deep learning techniques, employing various model architectures, have been extensively applied. Examples include recurrent neural networks (RNNs)\cite{NEURIPS2018_734e6bfc}, generative adversarial networks (GANs) \cite{luo2019e2gan}, and denoising diffusion probabilistic models \cite{tashiro2021csdi}. Although  these methods could  theoretically be applied  to soft sensing, their ability   to leverage  relationships between sensor measurements is often limited   by the absence of an  explicit representation of complex sensor measurement interactions. Graph Neural Networks (GNNs) have shown promise in effectively capturing intricate interactions and interdependencies among nodes in sensor networks \cite{zhou2020graph, wu21graph}. Generally, GNNs offer advantages in processing  multivariate time series data by forming a graph structure, where each time series is represented as a node, and the edges reflect the interactions between different time series \cite{jin2023survey}. For example, the study in \cite{cini2022filling} demonstrates the effectiveness of GNNs in imputing missing data, resulting in a possible application  of these capabilities to soft sensing \cite{felice2024graphbased, zhao2024virtual}. Moreover, within IIoT systems, GNNs have been successful in  capturing evolving relationships among multivariate time series data, which has led for example to  state-of-the-art results in early fault detection \cite{zhaodyedgegat2024, deng2021graph}. 

Despite the promising results achieved by GNNs in various tasks involving multivariate time series, several research gaps remain. Firstly, in scenarios  where sensors are sparsely distributed and measurements are scarce, there may not be enough  information for GNNs to extract the informative features necessary for downstream tasks. Furthermore, when GNNs are applied in a purely data-driven manner, it poses challenges to interpretability, which is crucial for decision-making in high-stakes industrial processes. Additionally, purely data-driven GNNs may lose their generalizability when applied across diverse  operating conditions. Therefore, the integration of physics principles with GNNs is an area that requires  further exploration to overcome  these limitations and enhance the robustness and generalizability of soft sensors.  However,  effectively embedding the laws of physics into GNNs for soft sensing remains an open research question due to the limited research in this field. This gap  hinders  GNNs' ability to achieve desired levels of interpretability, accuracy, and generalization in soft sensing applications. 

To overcome these limitations, we propose a novel approach to integrate physics principles into GNNs by augmenting graph structures with additional nodes derived from physics-based equations. This augmentation aims to improve the interpretability and the accuracy of soft sensor estimation by providing vital information for inferring unobserved variables.

In this research, we evaluate the proposed methodology through  a case study of district heating networks. These networks comprise an insulated piping system designed for heat distribution, where hot water transports  the generated heat to various substations. At these substations, flow, pressure, and temperature are regulated by pumps, valves, and control systems. Positioned to play a crucial role in the ongoing transformation of the energy sector, these systems represent scenarios with limited sensing capabilities \cite{lund20144th, van2023intelligent}. Recent studies have highlighted that many district heating networks suffer from suboptimal performance due to faults, leading to compromised consumer comfort \cite{maansson2019faults, ostergaard2022low}. Detecting and diagnosing faults in these networks presents challenges, as routine maintenance is not feasible  due to limited accessibility to the buried pipes, which are common in district heating networks.  Therefore, several methods have been developed to detect and diagnose faults in district heating networks using real-time condition monitoring data \cite{xue2020machine, wang2021fault}. However, given the limited number an types of available sensors, relying solely on these direct measurements may not be sufficient to detect faults using data-driven methods, highlighting the necessity for soft sensing. This limitation necessitates the development of algorithms that can effectively infer and estimate system parameters, leveraging the available data to ensure accurate monitoring in the absence of comprehensive sensor coverage. 

This study aims to estimate pressure and temperature values (considered as soft sensors)  across the entire district heating network using only the existing mass flow rate sensors. These pressure and temperature measurements are critical for ensuring the efficient and safe operation of district heating networks. They are regularly used  in monitoring and controlling system parameters, which leads to optimized network efficiency, ensured  safety of the infrastructure, and accurate billing of customers based on the actual  energy usage. We rely on characteristic parameters of the network's components, such as pipe diameter, length, roughness, and valve flow coefficient, to infer pressure and temperature. Therefore, we leverage these characteristic parameters to estimate virtual sensors beyond the raw sensor measurements obtained from mass flow rate sensors.  By incorporating this additional physics-based information derived from Darcy-Weisbach and heat transfer laws into the input graph as additional nodes, we introduce a physics-enhanced model for soft sensing. It is worth mentioning that graph data augmentation has proven effective for GNNs in downstream tasks \cite{liu2022local}. For instance, edge manipulation improves performance in semi-supervised node classification tasks \cite{zhao2021data}. Moreover, it has been shown that node augmentation based on convex combinations of existing nodes can generally enhance the GNN performance for downstream tasks  \cite{xue2021node}. However, our proposed physics-enhanced method augments  nodes by considering physics-based equations of the network, resulting in a more interpretable approach. It is worth mentioning that our proposed method for physics-based graph node augmentation can be applied to other types of networks. However, we chose to focus on district heating networks for several reasons. Firstly, this type of network has not been extensively studied using GNNs, presenting a unique opportunity to apply and evaluate our method in a relatively unexplored area. Additionally, district heating networks are inherently more complex than other pipeline networks, such as water distribution networks, due to the need to monitor both temperature and pressure. Furthermore, district heating networks are well-documented  in terms of physics and fluid dynamics, providing well-defined equations  that can be effectively integrated into GNN-based approaches.  

Finally, our experimental results demonstrate that using physics-enhanced GNNs can significantly improve performance, even in scenarios where physical sensor readings are susceptible to noise and physical parameters such as pipe roughness are subject to inaccuracy. 

The main contributions of this article are summarized as follows:

\begin{enumerate}


\item We propose a novel soft sensing model, based on physics-enhanced GNNs by creating new graph structures with augmented nodes that are derived from physics-based equations. 

\item We conduct experiments to evaluate the performance of the proposed method across different scenarios in a case study of a district heating network. Our results demonstrate that using physics-enhanced GNNs significantly improves performance and reduces  estimation error.

\item We have created a new dataset for district heating networks, which will be made openly available to support future studies on developing data-driven approaches for these networks.
\end{enumerate} 

The remainder of the paper is organized as follows: In Section \ref{sec:preliminaries}, we provide a comprehensive review of related backgrounds in graph signal processing and graph neural networks. Section \ref{sec:method} introduces the proposed physics-enhanced GNN methodology and details the incorporation of physics-based equations. Section \ref{sec:result} presents the experimental setup and results, showcasing the performance of our approach compared to several baseline methods. Finally, Section \ref{sec:conclusion} concludes the paper with a summary of our findings and outlines directions for future research in this area.

\section{Preliminaries}
\label{sec:preliminaries}

In this section, we start by reviewing some fundamental concepts in graph signal processing (GSP) \cite{ortega18gsp, shuman2013emerging}. Then, we introduce the background of two types of graph convolutional neural networks (GCNs) that have been widely used recently: spectral-based GCNs and spatial-based GCNs.

\subsection{Graph Signal Processing}
Let $\mathcal{G} = (\mathcal{V}, \mathcal{E}, \mathbf{A})$ be a weighted undirected graph, where $\mathcal{V}$, $\mathcal{E}$, and $\mathbf{A}$ represent the sets of nodes (vertices), edges, and the adjacency matrix, respectively. The topology of the graph is defined by its adjacency matrix $\mathbf{A}$ of size ${N \times N}$, such that the element \(\mathbf{A}(i,j)\) within this matrix indicates the edge weight between vertices \(v_i\) and \(v_j\). When no edge exists between \(v_i\) and \(v_j\), \(\mathbf{A}(i,j)\) is equal to zero. In the context of district heating networks, each sensor corresponds to a node within the graph $\mathcal{G}$, and the edges represent the similarity (e.g., Gaussian radial basis function) between these sensors. 

The normalized Laplacian matrix $\mathbf{L}$ is defined as \(\mathbf{L} := \mathbf{I_N} -\mathbf{D}^{-(1/2)} \mathbf{A}\mathbf{D}^{-(1/2)}\), where $\mathbf{D} = \mathsf{diag}(\mathbf{A1})$ is the diagonal matrix that contains information about the degree of each node, and $\mathbf{1}$ is a constant vector of ones. Since the normalized graph Laplacian matrix is positive semidefinite, by eigendecomposition, we have \(\mathbf{L} = \mathbf{U} \mathbf{\Lambda} \mathbf{U}^\mathsf{T}\), where \(\mathbf{U}\) is a unitary matrix of eigenvectors, and \(\mathbf{\Lambda} \in \mathbb{R}^{N \times N}\) denotes the diagonal matrix of corresponding eigenvalues. The graph Fourier transform (GFT) of the signal \(\mathbf{x}\) ($\mathcal{F}(\mathbf{x}) = \tilde{\mathbf{x}}$) is computed as \(\tilde{\mathbf{x}} = \mathbf{U}^\mathsf{T}\mathbf{x}\), with its inverse (IGFT) defined as \(\mathbf{x} = \mathbf{U} \tilde{\mathbf{x}}\). 

\subsection{Spectral-Based Graph Convolutional Neural Networks}
Spectral-based techniques,  which are grounded in the graph Fourier domain, have a strong theoretical foundation. These methods are typically applied to undirected graphs. In these methods, the convolution of an input signal $\mathbf{x}$ with a real signal $\mathbf{g} \in \mathbb{R}^n$ is described as follows:

\begin{equation}
\mathbf{x} *_{\mathcal{G}} \mathbf{g} = \mathcal{F}^{-1}(\mathcal{F}(\mathbf{x}) \odot \mathcal{F}(\mathbf{g})) = \mathbf{U}(\mathbf{U}^\mathsf{T} \mathbf{x} \odot \mathbf{U}^\mathsf{T} \mathbf{g}).
\end{equation}

Here, \(\odot\) represents the element-wise (Hadamard) product operator. If we denote a filter as \(\mathbf{g}_{\theta} = \mathsf{diag}(\mathbf{U}^\mathsf{T}  \mathbf{g})\), the spectral graph convolution can be expressed  as:

\begin{equation}
\mathbf{x} *_{\mathcal{G}} \mathbf{g}_{\theta} = \mathbf{U} \mathbf{g}_{\theta} \mathbf{U}^\mathsf{T} \mathbf{x}.
\end{equation}

Spectral-Based Graph Convolutional Neural Networks assume that the filter \( \mathbf{g}_{\theta} \) is a set of learnable parameters. However, the eigen-decomposition required for this approach has a computational complexity of \( \mathcal{O}(N^3) \), which is expensive for large-scale graphs. For this reason, some approaches such as Chebyshev Spectral Graph Convolutional Operator (ChebyNet) \cite{defferrard2016convolutional} and Fourier Graph Operator (FGO) \cite{yi2024fouriergnn} are proposed to reduce the computational complexity.



\subsection{Spatial-Based Graph Convolutional Neural Networks}
In contrast to spectral-based techniques that perform convolutions in the Fourier domain, spatial-based approaches directly establish convolution processes within the graph domain utilizing node information from nearby neighborhoods. Spatial-Based Graph Convolutional Neural Networks use the idea of message passing as follows:

\begin{equation}
\mathbf{m}_{\mathcal{N}}^{(k)}(u) = \mathsf{AGG} \left\{ \left(\mathsf{MSG}^{(k)} (\mathbf{h}_{u}^{(k)}, \mathbf{h}_{v}^{(k)}), \, \forall v \in \mathcal{N}(u) \right) \right\},
\end{equation}
\begin{equation}
\mathbf{h}_{u}^{(k+1)} = \mathsf{UPDATE}^{(k)} \left( \mathbf{h}_{u}^{(k)}, \, \mathbf{m}_{\mathcal{N}}^{(k)}(u) \right),
\end{equation}
where $\mathbf{h}_{u}^{(k)}$ represents the embedding of node $u$ at the $k$-th layer. The functions responsible for updating ($\mathsf{UPDATE}$) and messaging ($\mathsf{MSG}$) can be realized through any differentiable function, such as a multilayer perceptron (MLP). The aggregation operation ($\mathsf{AGG}$) represents a permutation-invariant aggregation operator. Here, $\mathcal{N}(u)$ denotes the set of neighboring nodes of node $u$. To improve the aggregation layer in GNNs, Graph Attention Networks (GATs) \cite{veličković2018graph, brody2022how} and Graph Transformer \cite{ijcai2021p214} are proposed to assign attention weights to each neighbor, regulating their influence during the aggregation process.

\begin{figure*}
\centerline{\includegraphics[width=\linewidth]{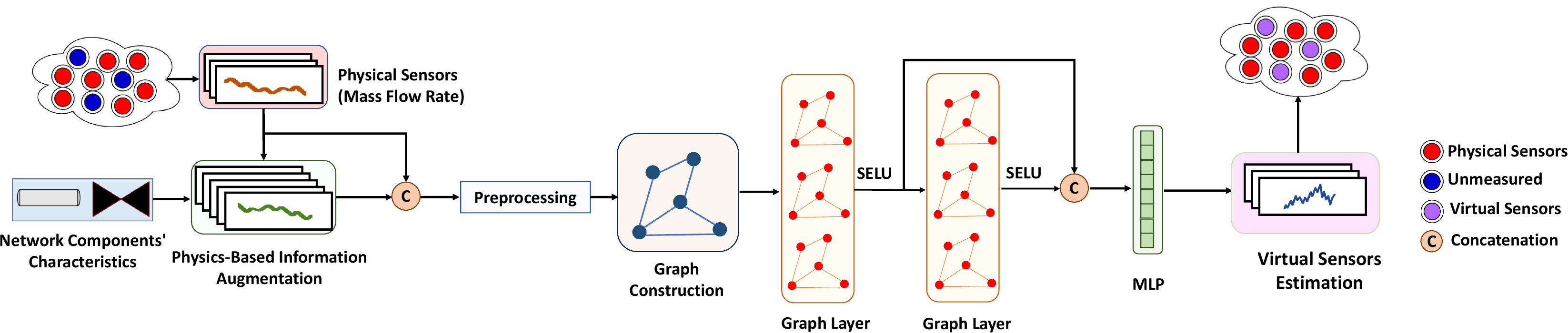}}
\caption{Overview of the proposed Physics-Enhanced GNN architecture. In the absence of measurements in certain areas, a subset of available physical sensors is utilized to compute supplementary physical information based on component characteristics. After the preprocessing step, a graph using both physical sensor data and the calculated information is constructed. This graph is then fed into two graph layers with a skip connection, followed by an MLP for virtual sensor estimation.}
\label{fig2}
\end{figure*}

\begin{figure*}
\centerline{\includegraphics[width=\linewidth]{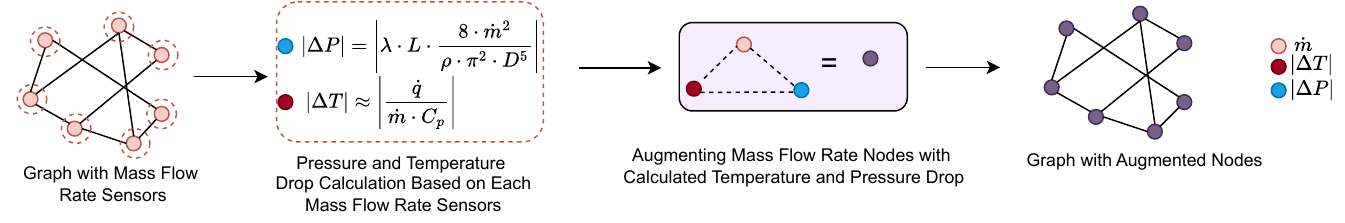}}
\caption{Graphs with augmented nodes, encompassing both physical mass flow rate sensors and physics-based information}
\label{augmented}
\end{figure*}

\section{Proposed Method}
\label{sec:method}

In this section, we will begin by explaining the general framework detailing how we integrate the characteristics of physical processes into GNNs (Section \ref{subsec:fusing}). Subsequently, we will delve deeper into the district heating network, which possesses unique physical process characteristics (Section \ref{subsec:physics-based info}). Within this subsection, we explore methods for extracting additional information based on the installed sensors and the characteristics of the physical process. Lastly, we explain the construction of the graph incorporating augmented physics-based nodes and mass flow rate sensors, designed for input into GNNs.

\subsection{Fusing Physics-Based and Data-Driven Methods}
\label{subsec:fusing}

In this study, we propose enhancing the input graph to the GNN with additional augmented nodes derived from physics-based equations. This enhancement allows us to improve the accuracy of soft sensor estimation, expanding the information available for constructing data-driven soft sensor models. While several approaches have previously been proposed to incorporate physics-based information into data-driven models, our proposed approach focuses on adding additional relevant variables to the data-driven estimators' input. Figure \ref{fig2} illustrates the framework and architecture of our proposed physics-enhanced soft sensing approach. In contrast to purely data-driven approaches, which aim to learn a mapping function solely from mass flow rate sensors ($\dot{m}$) to the temperature and pressure targets $\mathbf{y}$ (i.e., $\dot{m} \rightarrow \mathbf{y}$), our framework takes the graph with mass flow rate measurements as inputs, combined with physical process stem from components characteristics parameters.  In the initial stage, these parameters are employed to calculate physics-based information. Subsequently, they are combined with sensors measuring physical mass flow rates in the second step leading to the construction of a larger graph with more nodes, encompassing both physical mass flow rate sensors and physics-based information as illustrated in Figure \ref{augmented}. As illustrated in the Figure, the initial graph is comprised solely of mass flow rate sensors. In our proposed approach, we augment the graph by incorporating specific mass flow rates of water within the pipes, along with network component characteristics such as pipe length ($L$), pipe diameter ($D$), and pipe cross-sectional area. Subsequently, we compute the pressure drop ($\Delta P$) and temperature drop ($\Delta T$) in the water as it moves through the pipes, using established fluid dynamics equations. These computed values are then added as additional nodes, thereby enriching the data representation and providing a more detailed and informative model. Our proposed framework is highly adaptable and can be integrated with various physics-based methods and architectures.

\subsection{Physics-Based Information Augmentation}
\label{subsec:physics-based info}
The primary objective of this subsection is to introduce the methodology for effective physics-based information augmentation (temperature and pressure variations within networks)  based on the physical characteristics of components and the available measurements. These variations in pressure and temperature within the network are closely related to the dynamics of fluid flow through the pipes. To establish an effective approach for implementing GNN-based soft sensing, it is advantageous to begin by examining the underlying physics and mathematical principles of district heating networks.  In particular, the governing equations that describe fluid flow in a pipe network include the continuity equation, which can be expressed  as follows:

\begin{equation}
 \sum_i \dot{m}_{in,i} - \sum_o \dot{m}_{out,o} = 0,
\end{equation}
where
$\dot{m}_{in,i}$ denotes the mass flow rate ($\mathrm{kg/s}$) entering the system at inlet $i$, and
$\dot{m}_{out,o}$ represents the mass flow rate exiting the system at outlet $o$. Furthermore, according to Bernoulli's principle, the flow speed of a fluid increases as its pressure or potential energy drops. In other words, the sum of a fluid's potential energy, pressure, and speed remains constant. Therefore, for the pipe illustrated in Fig.\ \ref{fig1}, we can express  Bernoulli's equation as follows: 

\begin{figure}
\centerline{\includegraphics[width=\linewidth]{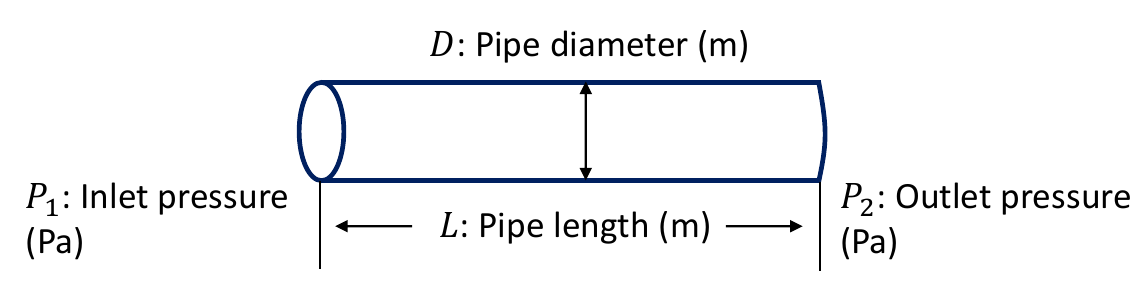}}
\caption{Pipe with its physical parameters.}
\label{fig1}
\end{figure}

\begin{equation}
P_{1}+\frac{1}{2}\rho v_{1}^{2}+\rho gh_{1}=P_{2}+\frac{1}{2}\rho v_{2}^{2}+\rho gh_{2}+H_{F},
\end{equation}
where $P_1$ and $P_2$ represent the inlet and outlet pressures (in $\mathrm{Pa}$), respectively, $v$ denotes  the fluid velocity ($\mathrm{m/s}$), $\rho$  stands for the density of water ($\mathrm{kg/m^3}$), $g$ represents the gravitational acceleration ($\mathrm{m/s^2}$), $h$ is the height ($\mathrm{m}$), and $H_{F}$ represents  the head loss due to friction, which can be calculated using the Darcy-Weisbach equation:

\begin{equation}
H_{F} = \frac{\rho \cdot v^2 \cdot  \lambda(\text{Re}, k_s, D) \cdot L}{2 \cdot D},
\end{equation}
where $L$ represent the pipe length ($\mathrm{m}$), $\lambda$ stands for the Darcy friction factor, which itself depends on the pipe’s roughness $k_s$, diameter $D$, and the Reynolds number $\text{Re}$. 

In practical scenarios,  Darcy friction factors can be determined using the Moody diagram or other methods. For laminar flow, it can be calculated as $\lambda = 64/\text{Re}$. However, for turbulent flow,  the Colebrook-White equation can be used in a recursive manner:

\begin{equation}
\frac{1}{\sqrt{\lambda}} = -2 \cdot \log_{10}\left(\frac{k_s}{3.7 \cdot D} + \frac{2.51}{{\rm Re} \cdot \sqrt{\lambda}}\right)
\end{equation}
Assuming that the pipe is horizontal ($h_1 = h_2$) and has a constant diameter ($v_1 = v_2$), we have $
|\Delta P| = |H_F|
$. Then, for a cylindrical pipe:

\begin{equation}
|\Delta P| = \left|\lambda \cdot L \cdot \frac{8 \cdot \rho \cdot Q^2}{\pi^2 \cdot D^5}\right|,
\end{equation}
where $Q$ represents  the volumetric flow rate and $\dot{m} = \rho Q$.
Hence, we have:

\begin{equation}
|\Delta P| = \left|\lambda \cdot L \cdot \frac{8  \cdot \dot{m}^2}{\rho \cdot \pi^2 \cdot  D^5}\right|
\end{equation}
In this equation, while the sign of $\Delta P$ can indeed be inferred from the direction of the mass flow rate, our research specifically emphasizes the magnitude of the pressure drop.

Another alternative is the Hazen-Williams equation:

\begin{equation}
|\Delta P| = \left|\frac{10.67 \cdot L \cdot Q^{1.852}}{C^{1.852} \cdot D^{4.87}}\right|,
\end{equation}
where $C$ represents the Hazen-Williams roughness coefficient (dimensionless). It offers a simple and practical method for estimating pressure loss in water pipes due to friction, eliminating the need to calculate the friction factor. However, it may not be as accurate as the Darcy-Weisbach equation. In our study, we opted for the Darcy-Weisbach equation since we had all the necessary physical parameters. Nevertheless, we will demonstrate that even when precise parameters are unavailable, our physics-based augmented information can still contribute effectively to soft sensing. Furthermore, valves have a crucial role in district heating networks as they regulate and control water flow. Valves are responsible for maintaining pressure and ensuring a steady supply throughout the distribution system. The pressure difference caused by a valve can be calculated using the following equation:

\begin{equation}
|\Delta P| = \left|\frac{ G \cdot \dot{m}^2}{(\rho \cdot \zeta)^2}\right|,
\end{equation}
where $G \approx 1$ represents the specific gravity of water, and $\zeta$ is the valve flow coefficient, which is related to the geometry of the valve.

For well-insulated pipes, which are commonly used in district heating networks, the temperature difference ($\Delta T$) can be approximated as:

\begin{equation}
|\Delta T| \approx \left|\frac{\dot{q}}{\dot{m} \cdot C_p}\right|,
\end{equation}
where $\dot{q}$ represents the heat energy transfer rate ($\mathrm{kW}$), and $C_p$ is the specific heat capacity of water ($\mathrm{kJ\cdot kg^{-1}\cdot K^{-1}}$).

\subsection{Modeling Network Measurements as Graph Signals}

We represent a segment comprising mass flow rate sensor measurements and the derived augmented physical processes (magnitude of temperature and pressure differences) as signals on a graph. The edge weights \(\mathbf{A}(i,j)\) for each segment are calculated by applying a thresholded Gaussian kernel to the pairwise Euclidean distance between nodes \(v_i\) and \(v_j\), 

\begin{equation}
\mathbf{A}(i,j) =
\begin{cases}
\exp\left(-\frac{{\text{dist}(v_i,v_j)^2}}{{2\sigma_0^2}}\right) & \text{if } \text{dist}(v_i, v_j) \leq \kappa, \\
0 & \text{otherwise}
\end{cases},
\label{graphconstruction}
\end{equation}
Here, \(\sigma_0\) represents the standard deviation of the distances, and \(\kappa\) is a threshold to impose sparsity.

\section{Experimental Results}
\label{sec:result}

\subsection{Case Study}
\begin{figure}
\centerline{\includegraphics[width=\linewidth]{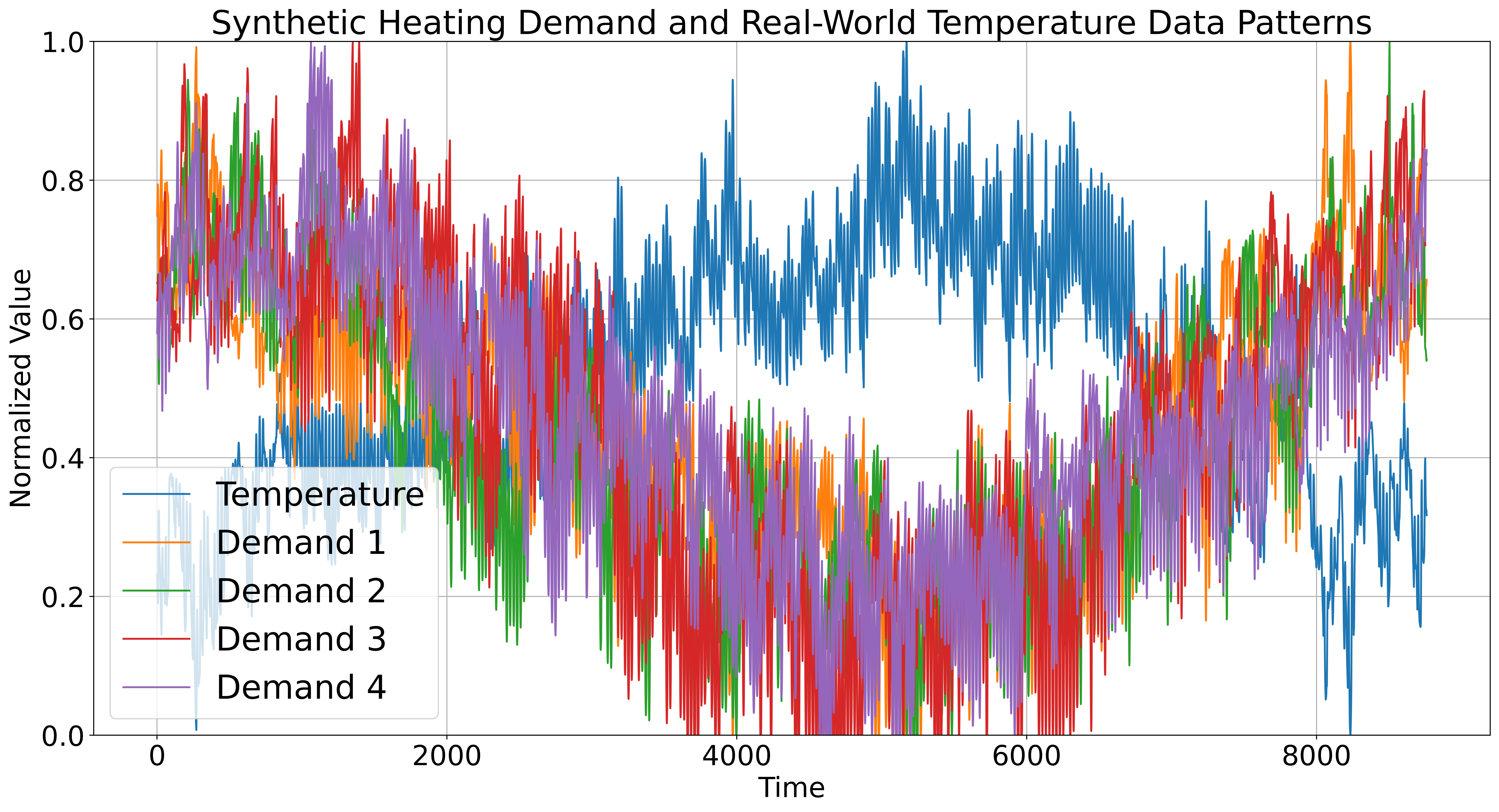}}
\caption{Heating demand and outside temperature patterns}
\label{heatingdemand}
\end{figure}

\begin{figure*}
\centerline{\includegraphics[width=\linewidth]{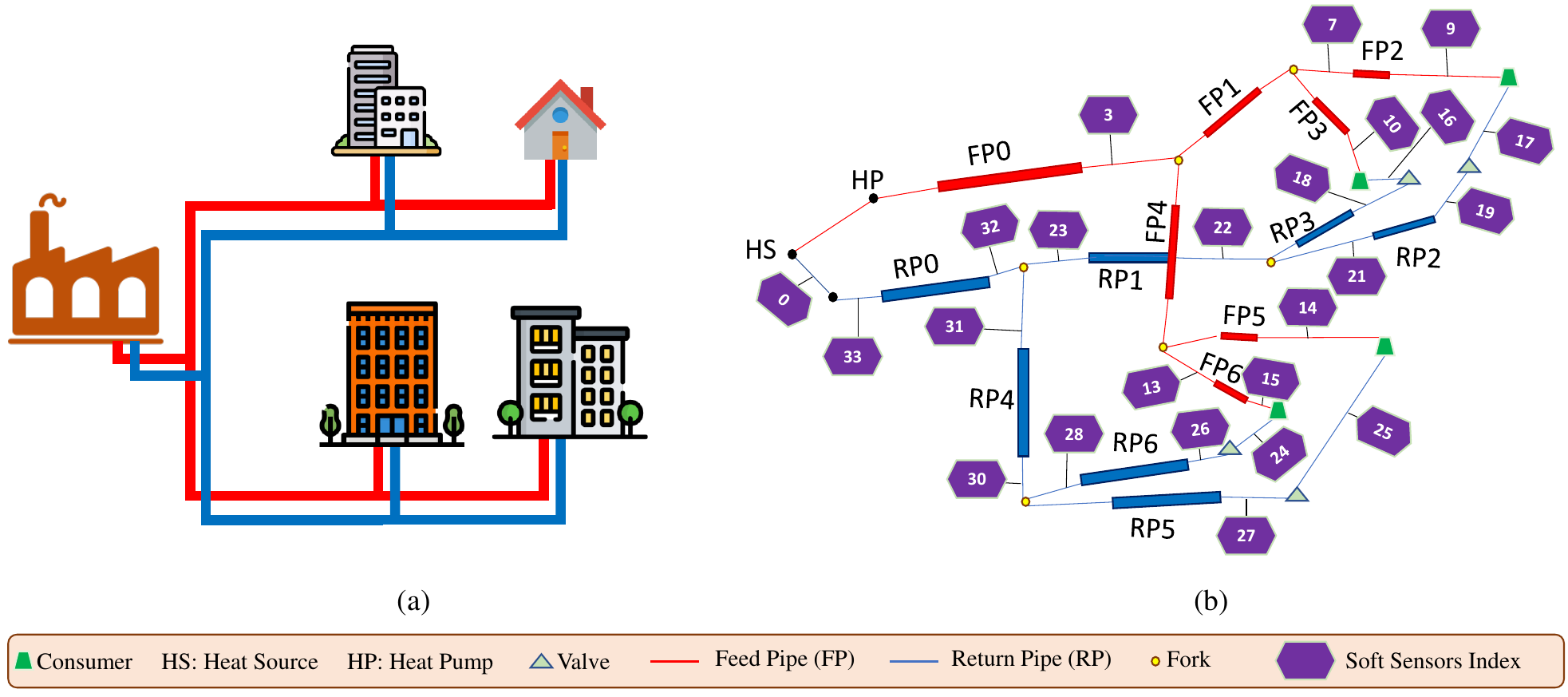}}
\caption{(a) Overview of the simulated district heating network, (b) Layout of the network simulation.}
\label{layout}
\end{figure*}

In this study, we conduct a case study using a simulation of a district heating network serving four consumers. Demand patterns for each consumer are generated based on one year of recorded weather temperature data with an hourly sampling rate (8760 samples for the entire year). These patterns have an inverse relationship with weather temperature and a direct correlation with the area of the building, indicating that heat demand decreases as weather temperature rises and increases with the size of the building as depicted in Fig.\ \ref{heatingdemand}. The complexity of district heating networks presents a challenge when studying them from a data-driven perspective, receiving comparatively less attention than other networks. This is primarily attributed to the scarcity of openly available datasets in these networks. While various networks such as water and electricity benefit from numerous benchmark datasets, supporting the application of data-driven approaches for tasks like anomaly detection \cite{cook2019anomaly, gardharsson2022graph}, state estimation \cite{tariq2020vulnerability}, soft sensing, and model calibration \cite{zanfei2023shall}, the limited availability of such datasets in district heating networks hinders the widespread use of data-driven methodologies. District heating networks comprise various components, including a heat source, heat pump, pipes, splitters and mergers (forks), and valves. We apply the TESPy library \cite{witte2020tespy} for steady-state simulation to capture the system's behavior as shown in Fig.\ \ref{layout}.  The water temperature in the pipes ranges from 50 to 90 degrees Celsius, depending on whether it is supplied or returned. Additionally, the water pressure in the pipes varies from 3 to 5 bar (300 to 500 kPa). Furthermore, Table \ref{tab:pipepara} provides detailed information on the physical attributes of the feed (FPs) and return pipes (RPs). We use PyTorch and PyTorch Geometrics \cite{Fey2019wv} for implementing neural networks and GNNs. To provide a comprehensive evaluation, we consider two different scenarios. The first scenario represents an ideal condition, devoid of any noise, allowing us to investigate the physics-based augmented nodes more accurately. In the second scenario, we introduce zero-mean Gaussian noise with a standard deviation of $0.1$ to the mass flow rate sensors and derive physics-based nodes. This approach accounts for potential variations in physical parameters over time, such as pipe roughness, inaccuracies in heat transfer rate, or reductions in sensor accuracy. 

\begin{table}[h]
\caption{Physical Characteristics of the Pipes}
\centering
\resizebox{0.85\linewidth}{!}{%
\begin{tabular}{ccccc}
\toprule
Pipe ID & $D~(\mathrm{mm})$ & $L~(\mathrm{m})$ & $k_s~(\mathrm 1/ m)$ & $k_a~(\text{W/\text{K}})$ \\
\midrule
FP0 & 31.3 & 60 & 0.0005 & 3.02 \\
FP1 & 27.6 & 120 & 0.0005 & 3.03 \\
FP2 & 17.5 & 20 & 0.0005 & 2.81 \\
FP3 & 16.9 & 100 & 0.0005 & 3.07 \\
FP4 & 27.6 & 120 & 0.0005 & 3.03 \\
FP5 & 16.7 & 20 & 0.0005 & 2.81 \\
FP6 & 18.5& 100 & 0.0005 & 3.06 \\
RP0 & 18.2 & 60 & 0.0005 & 3.94 \\
RP1 & 30.3 & 120 & 0.0005 & 3.99 \\
RP2 & 19.0 & 20 & 0.0005 & 3.83 \\
RP3 & 18.5 & 100 & 0.0005 & 4.29 \\
RP4 & 30.3 & 120 & 0.0005 & 3.85 \\
RP5 & 17.3 & 20 & 0.0005 & 3.68 \\
RP6 & 19.3 & 100 & 0.0005 & 4.04 \\
\bottomrule
\end{tabular}%
}
\label{tab:pipepara}
\vspace{0.2cm}

$D$: Pipe diameter, $L$: Pipe length, $k_s$: Pipe roughness, $k_a$: Area independent heat transfer coefficient
\end{table}

\subsection{Networks Implementation and Evaluation}
We evaluate different types of architecture for graph layers, including ChebyNet, GATv2, FGO, and Graph Transformer, as illustrated in Fig\@.  \ref{fig2}. After augmenting physics-based nodes and constructing the graph, we pass the resulting graph through two graph layers to obtain new node representations.  Throughout the training process, we use the mean squared error loss function and the Nesterov accelerated Adam (NAdam) optimizer for updating the weights. The hyperparameters were fine-tuned through a grid search over specific intervals, with parameter selection guided by the optimization of the validation loss. More precisely, the grid search spanned the interval $[0, 8]$ for the number of heads in the graph attention layer. Additionally, we explored values for the Chebyshev filter order within the range $[1, 8]$. The window size underwent investigation within the intervals $\{2, 4, 8, 12, 16\}$, while the hidden dimensions of the graph layer were searched within the set $\{8, 16, 32, 64\}$. Moreover, the hidden dimension of the multilayer perceptron (MLP) was explored across $\{32, 64, 128, 256\}$. It was observed that incorporating skip connections between graph layers led to improved performance, aligning with prior research indicating potential oversmoothing issues with an increased number of layers in GNNs \cite{oversmoothi}. Finally, the outputs from the two graph layers were concatenated and fed into an MLP for virtual sensor estimation. It is worth mentioning that the Rectified Linear Unit (ReLU) activation function is not well-suited for our case, particularly when a unit remains unactivated by any input within the dataset. This scenario poses a challenge for the gradient-based optimization algorithm, as it struggles to update the weights of the inactive unit. To address this issue, we employed the Scaled Exponential Linear Units (SELU) activation function after each layer. SELU enhances the Rectified Linear Unit (ReLU) by adding a slope on the negative half-axis. Table~\ref{tab1} provides a comprehensive overview of the optimal hyperparameter settings for our experiments.

\begin{table}[h!]
\centering
\caption{Hyperparameters Used in the Experiment}

\begin{tabular}{l l}

 \toprule
 Hyperparameter & Value \\
 \midrule\midrule
 Number of multi-head attention & $5$  \\ 
 Chebyshev order & $4$ \\
 $\kappa$ & $1$  \\
 Window size & $8$ \\
 Window stride & $1$ \\
 Hidden dimension (graph layer) & $16$  \\
 Hidden dimension (MLP) & $128$  \\
Optimizer & NAdam  \\ 
Learning rate & $3 \times 10^{-4}$ \\
Activation function & SELU \\
Patience for early stopping & $200$ \\
Batch size & $64$ \\
 \bottomrule
\end{tabular}
\label{tab1}
\end{table}

\begin{figure*}

\includegraphics[width=\linewidth]{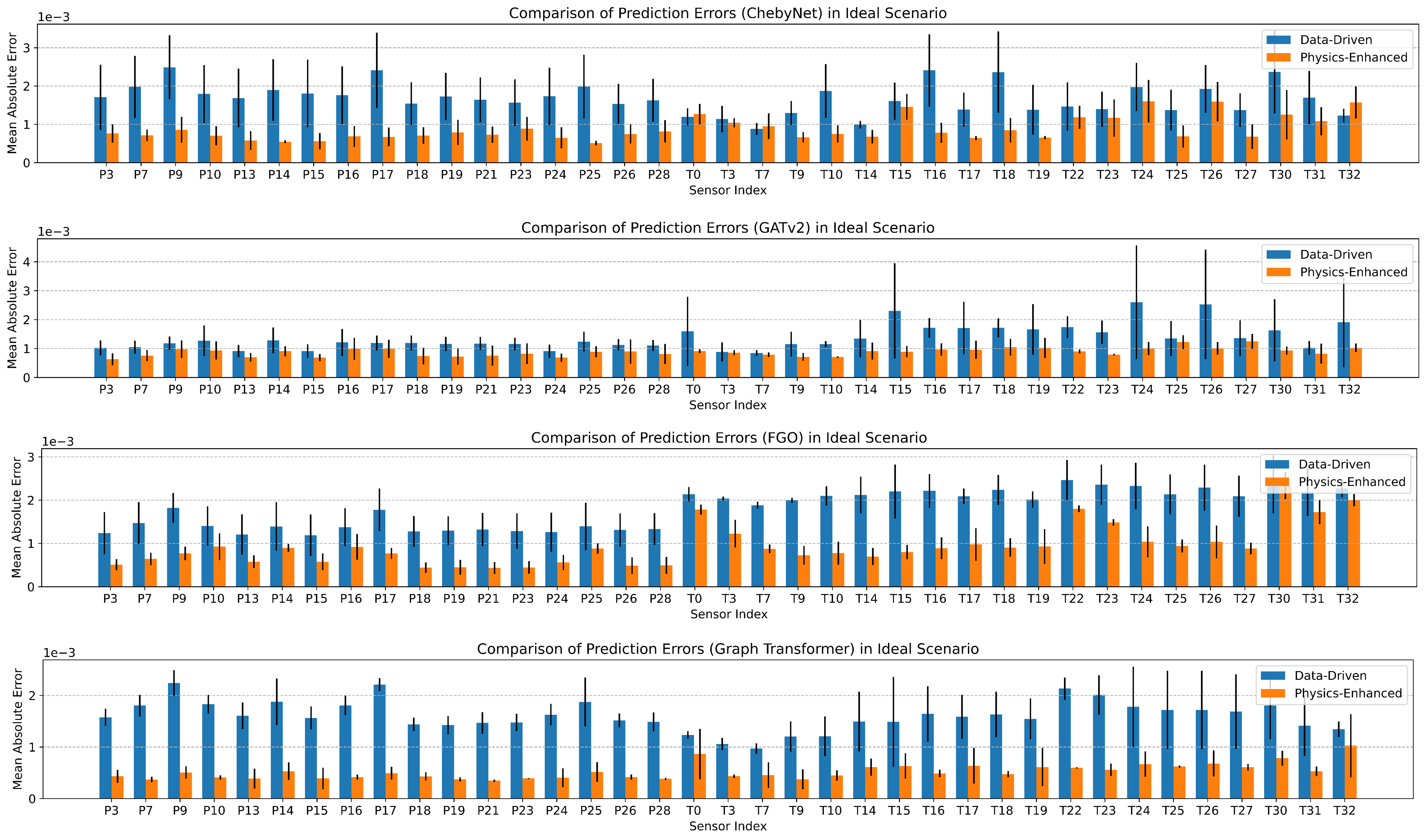}

\caption{Comparison of Sensor-Wise Mean Absolute Errors (MAE) for graph-based models in ideal scenario.}
\label{combined errorbar}
\end{figure*}

\begin{figure*}

\includegraphics[width=\linewidth]{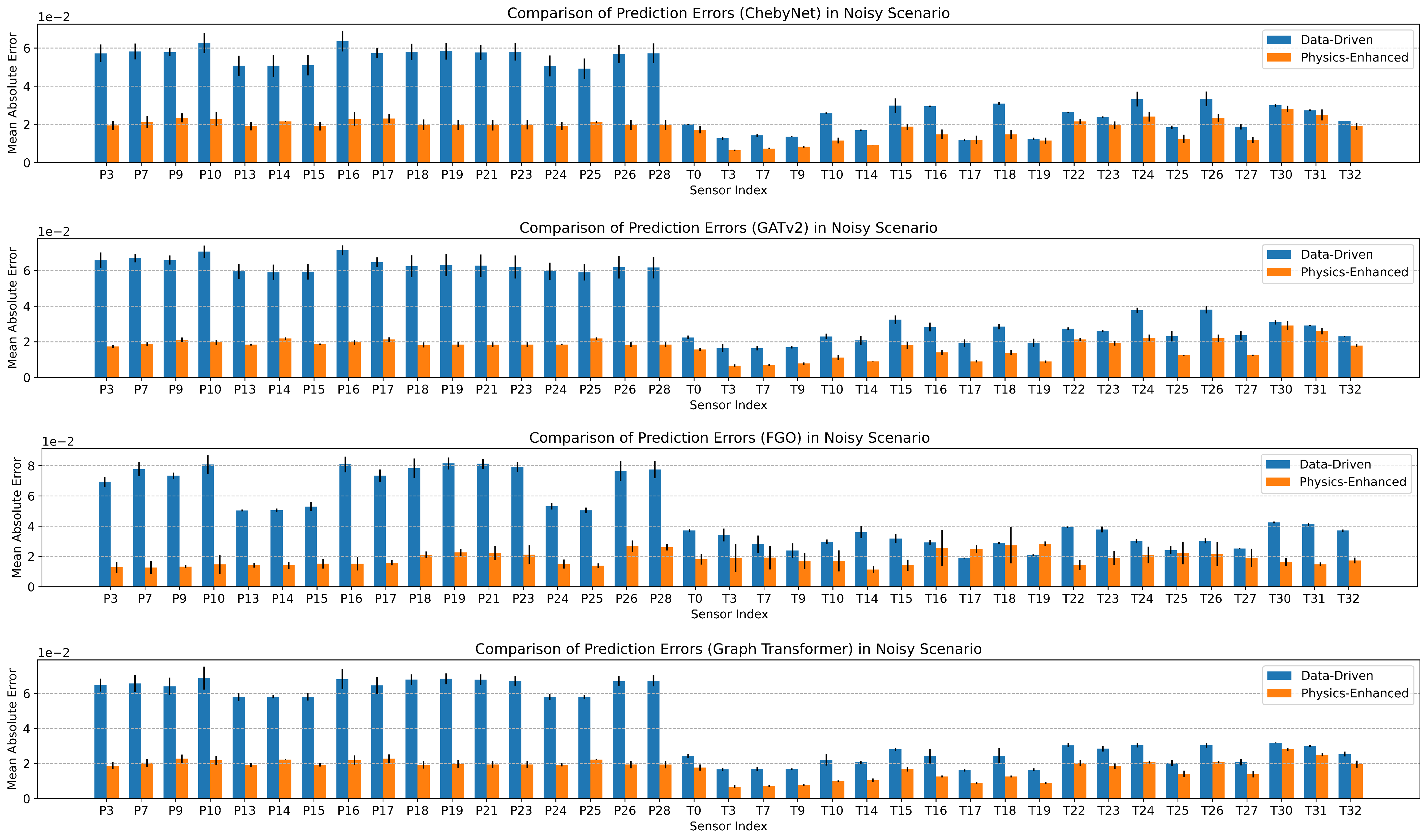}

\caption{Comparison of Sensor-Wise Mean Absolute Errors (MAE) for graph-based models in noisy scenario.}
\label{errorbar_noisy}
\end{figure*}

\subsection{Data Preprocessing}
For training and evaluation, we split the dataset based on its chronological sequence to maintain the temporal order of the data. This approach allows us to test and train the model on various time periods while ensuring its robustness and generalizability. Initially, 80\% of the data was allocated for training, with the remaining 20\% evenly divided between validation and testing sets. To prevent overfitting, we implemented early stopping during training. Subsequently, we selected the best model, determined through the validation data, for testing. Min-max normalization is also applied based on the training dataset. Once the data were appropriately normalized, we segmented mass flow sensor measurements and derived physical processes into different samples using a sliding window approach. This segmentation is crucial for the temporal construction of the graph. For both training and validation, we used a batch size of 64 with block-diagonal batching, which is a common method for graph-structured data.



\subsection{Evaluation Metrics}

In the context of the soft sensing task, we evaluate and compare the performance of the proposed framework with purely data-driven approaches. We employ three commonly used regression evaluation metrics:  root-mean-square error (RMSE), mean absolute error (MAE), and accuracy which are defined as follows:

\begin{equation}
\mathsf{RMSE} = \sqrt{\frac{1}{n} \sum_{i=1}^{n}(y_i - \hat{y}_i)^2},
\end{equation}

\begin{equation}
\mathsf{MAE} = \frac{1}{n} \sum_{i=1}^{n} |y_i - \hat{y}_i|,
\end{equation}

\begin{equation}
\mathsf{Accuracy} = 1 - \frac{\|\mathbf{Y} - \hat{\mathbf{Y}}\|_F}{\|\mathbf{Y}\|_F},
\end{equation}

where the variables \(n\), \(y_i\), \(\hat{y}_i\), and $\bar{y}$ are the total number of samples, actual observed value for the \(i\)-th estimation, predicted value for the \(i\)-th estimation, and the mean of the observed data, respectively. $\mathbf{Y}$ and $\mathbf{\hat{Y}}$ denote the sets of $y$ and $\hat{y}$, respectively. Accuracy is a measure of prediction precision. Unlike RMSE and MAE, where smaller values are better, accuracy is considered better when it has larger values. It is important to emphasize that the soft sensors, namely temperature and pressure, operate within significantly different value ranges. Consequently, we assess the model's performance using normalized values to ensure fair and meaningful comparisons.

\begin{table}[h!]
\centering
\caption{Model Performance Comparison on the Ideal (Without Noise) Case with 3 Runs. The Best Metric Result is Highlighted in \textbf{Bold}, and the Second Best Model is \underline{Underlined}.}
\label{tab:model-comparison}
\resizebox{\columnwidth}{!}{
\begin{tabular}{cccc}
\toprule
\textbf{Model} & \textbf{RMSE ($\times 10^{-3}$)} & \textbf{MAE ($\times 10^{-3}$)} & \textbf{Accuracy ($\times 10^{-2}$)} \\
\midrule\midrule
\textbf{MLP} & & & \\\midrule
Data-Driven & 2.39 $\pm$ 0.19 & 1.70 $\pm$ 0.16 & 99.60 $\pm$ 0.03 \\
Physics-Enhanced & 1.83 $\pm$ 0.06 & 1.35 $\pm$ 0.07 & 99.69 $\pm$ 0.01 \\
rel. Delta & -23.43\% & -20.59\% & 0.09\% \\
\midrule\midrule
\textbf{CNN} & & & \\\midrule
Data-Driven & 2.83  $\pm$ 0.42 & 2.13  $\pm$  0.38 & 99.53 $\pm$ 0.69 \\
Physics-Enhanced & 2.07 $\pm$ 0.07  & 1.45 $\pm$ 0.05  & 99.65 $\pm$ 0.01  \\
rel. Delta & -26.86\%  & -31.92\%  &  0.12\% \\
\midrule\midrule
\textbf{ChebyNet} & & & \\\midrule
Data-Driven & 4.36 $\pm$ 2.37 & 1.68 $\pm$ 0.51 & 99.27 $\pm$ 0.40 \\
Physics-Enhanced & \underline{1.25} $\pm$ 0.06 & \underline{0.88} $\pm$ 0.08 & \underline{99.79} $\pm$ 0.01 \\
rel. Delta & -71.33\% & -47.62\% & 0.52\%  \\
\midrule\midrule
\textbf{GATv2} & & & \\ \midrule
Data-Driven & 2.13 $\pm$ 0.46 & 1.37 $\pm$ 0.38 & 99.64 $\pm$ 0.08 \\
Physics-Enhanced & 1.29 $\pm$ 0.10 & \underline{0.88} $\pm$ 0.12 & 99.78 $\pm$ 0.02 \\
rel. Delta & -39.44\% & -35.77\% & 0.14\%  \\
\midrule\midrule

\textbf{FGO} & & & \\
\midrule
Data-Driven & 3.55 $\pm$ 1.88 & 1.81 $\pm$ 0.32 & 99.41 $\pm$ 0.31 \\
Physics-Enhanced & 1.38 $\pm$ 0.18 & 0.92 $\pm$ 0.16 & 99.77 $\pm$ 0.04  \\
rel. Delta & -61.13 \% & -49.17 \% & 0.36\%  \\
\midrule\midrule

\textbf{Graph Transformer} & & & \\
\midrule
Data-Driven & 4.12 $\pm$ 1.7 & 1.61 $\pm$ 0.19 & 99.31 $\pm$ 0.28 \\
Physics-Enhanced & \textbf{0.77} $\pm$ 0.12 & \textbf{0.52} $\pm$ 0.10 & \textbf{99.87} $\pm$ 0.02 \\
rel. Delta & -81.31\% & -67.7\% & 0.56\%  \\
\bottomrule
\label{ideal}
\end{tabular}
}
\end{table}

\begin{figure}

\centering
\begin{minipage}[b]{\linewidth}
  \centering
  \includegraphics[width=\linewidth]{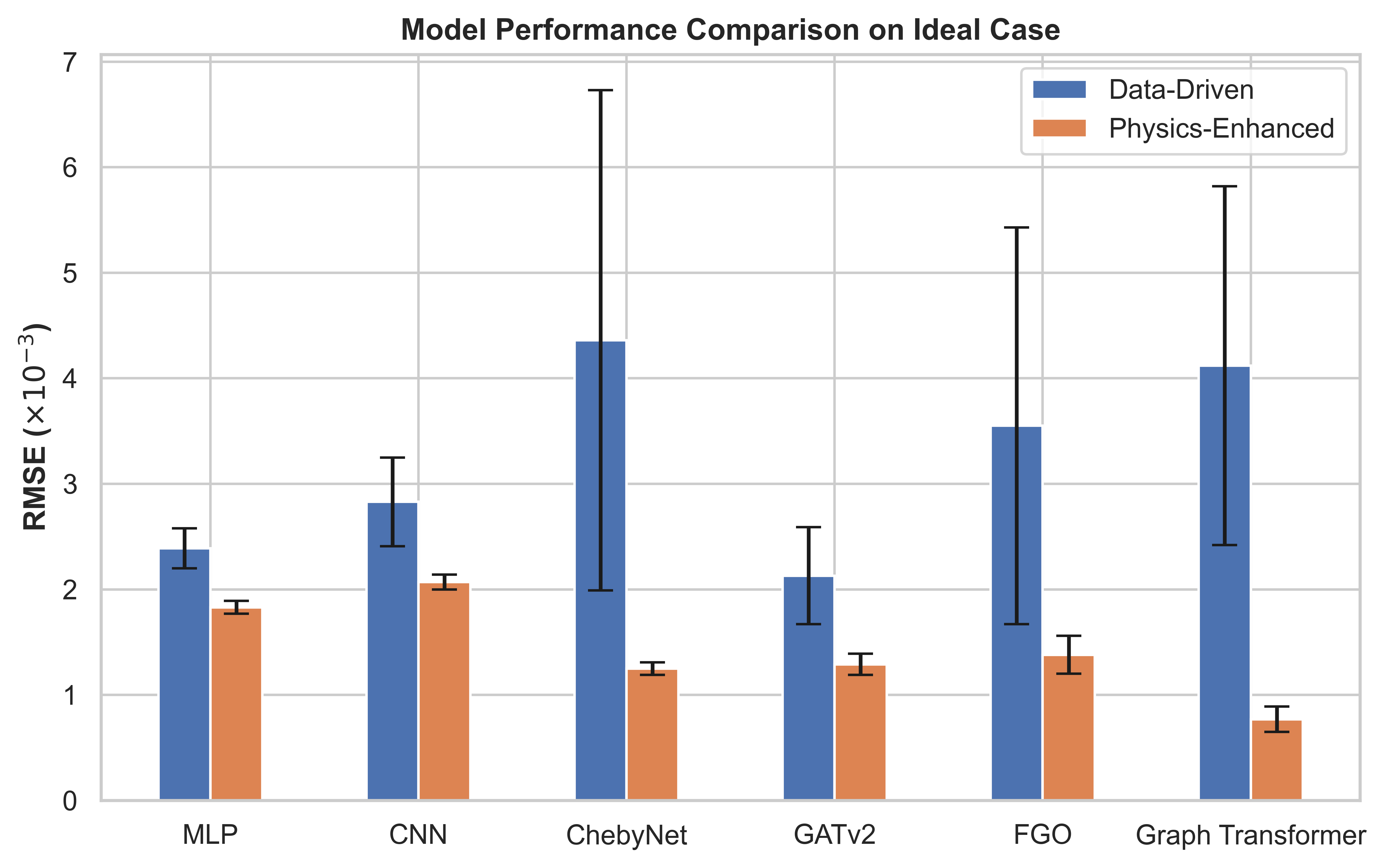}

\end{minipage}\qquad
\begin{minipage}[b]{\linewidth}
  \centering
  \includegraphics[width=\linewidth]{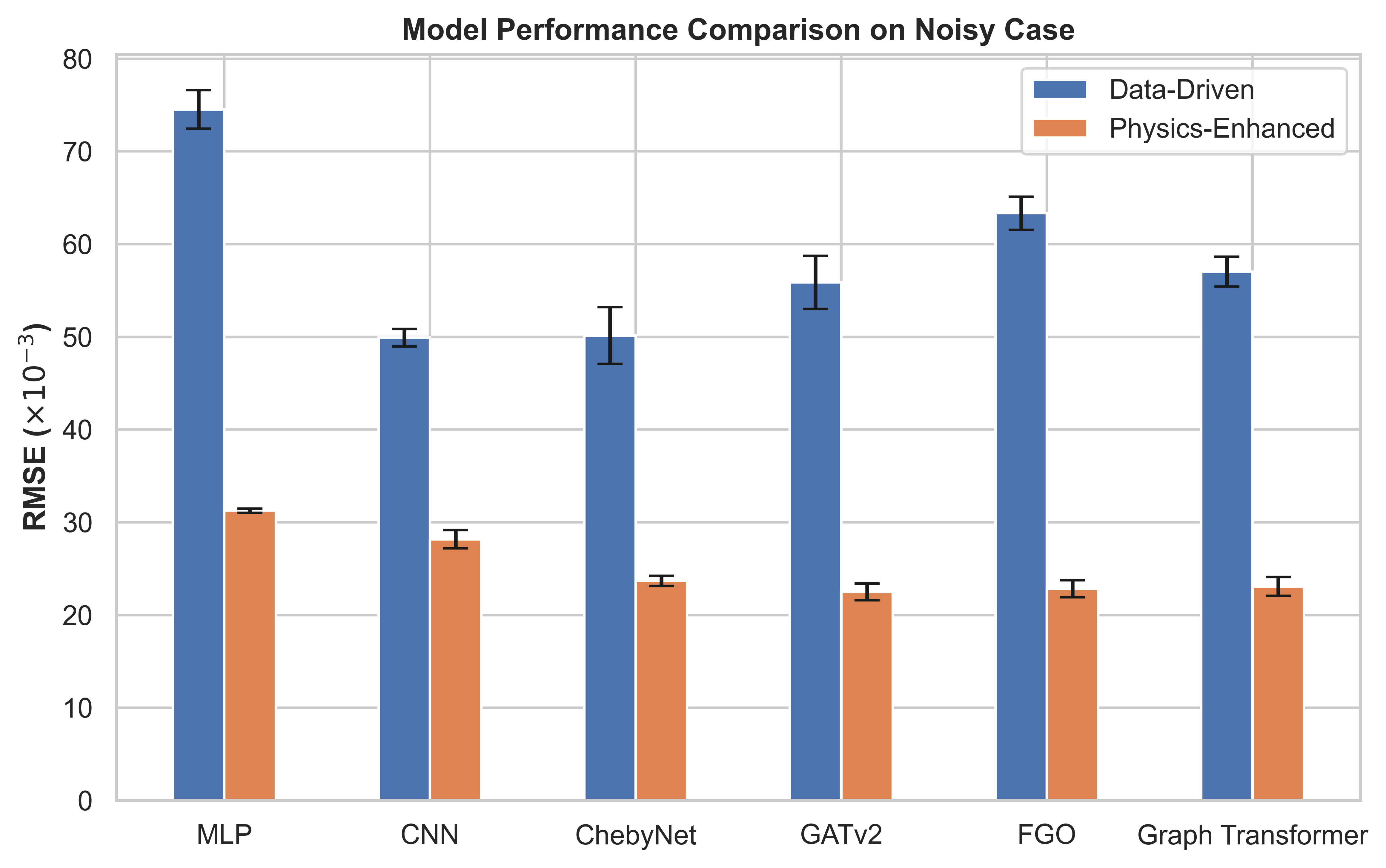}
 
\end{minipage}
\caption{Comparative Analysis of Model Performance in Ideal and Noisy Conditions.}
\label{fig: overalperformance}
\end{figure}

\subsection{Results and Discussion}

In this section, we compare the performance of the proposed physics-enhanced approach with the purely data-driven approach. It is worth mentioning that we consider both noise-free and noisy scenarios in our evaluation, justifying the robustness of the compared approaches using the aforementioned metrics. The proposed method for virtual sensor estimation is compatible with different neural networks. In our case study, we evaluate MLP, CNN, ChebyNet, GATv2, FGO, and Graph Transformer since they possess distinct characteristics. Here, we consider purely data-driven approaches as our baseline, while the physics-enhanced approach serves as our proposed method.

\begin{table}[h!]
\centering
\caption{Model Performance Comparison on the Noisy Case with 3 Runs. The Best Metric Result is Highlighted in \textbf{Bold}, and the Second Best Model is \underline{Underlined}.}
\label{tab:model-comparison}
\resizebox{\columnwidth}{!}{
\begin{tabular}{cccc}
\toprule
\textbf{Model} & \textbf{RMSE ($\times 10^{-3}$)} & \textbf{MAE ($\times 10^{-3}$)} & \textbf{{Accuracy ($\times 10^{-2}$)}} \\
\midrule\midrule
\textbf{MLP} & & & \\\midrule
Data-Driven & 74.53 $\pm$ 2.08 & 57.68 $\pm$ 1.36 & 87.54 $\pm$ 0.35 \\
Physics-Enhanced & 31.25 $\pm$ 0.22 & 23.77 $\pm$ 0.18 & 94.78 $\pm$ 0.04 \\
rel. Delta & -58.07\% & -58.79\% & 8.27\% \\
\midrule\midrule
\textbf{CNN} & & & \\\midrule
Data-Driven & 49.91  $\pm$ 0.95 & 37.26 $\pm$ 0.80  & 91.66  $\pm$ 0.16  \\
Physics-Enhanced & 28.17  $\pm$ 0.98 & 21.61 $\pm$ 0.74  & 95.29 $\pm$ 0.16\\
rel. Delta & -43.55\%  & -42.00\%   &  3.96\%  \\
\midrule\midrule
\textbf{ChebyNet} & & & \\\midrule
Data-Driven & 50.14 $\pm$ 3.06 & 38.01 $\pm$ 2.53 & 91.62 $\pm$ 0.51 \\
Physics-Enhanced & 23.69 $\pm$ 0.54 & 18.06 $\pm$ 0.41 & 96.04 $\pm$ 0.09 \\
rel. Delta & -52.75\% & -52.49\% & 4.82\%  \\
\midrule\midrule
\textbf{GATv2} & & & \\ \midrule
Data-Driven & 55.87 $\pm$ 2.87 & 42.66 $\pm$ 2.39 & 90.66 $\pm$ 0.48 \\
Physics-enhanced & \textbf{22.50} $\pm$ 0.90 & \textbf{17.05} $\pm$ 0.71 & \textbf{96.24} $\pm$ 0.15 \\
rel. Delta & -59.73\% & -60.03\% & 6.16\%  \\
\midrule\midrule

\textbf{FGO} & & & \\
\midrule
Data-Driven & 63.34 $\pm$ 1.78 & 49.05 $\pm$ 1.16 & 89.41 $\pm$ 0.29 \\
Physics-Enhanced & \underline{22.84} $\pm$ 0.91 & 18.51 $\pm$ 0.99 & \underline{96.18} $\pm$ 0.15  \\
rel. Delta & -63.94 \% & -62.26\% & 7.57\%  \\
\midrule\midrule

\textbf{Graph Transformer} & & & \\
\midrule
Data-Driven & 57.05 $\pm$ 1.61 & 42.35 $\pm$ 1.71 & 90.46 $\pm$ 0.27 \\
Physics-Enhanced & 23.09$\pm$ 1.00 & \underline{17.53} $\pm$ 0.79 & 96.14 $\pm$ 0.16 \\
rel. Delta & -59.53\% & -58.61\% & 6.28\%  \\
\bottomrule
\label{noisy}
\end{tabular}
}
\end{table}

\begin{figure}
\centerline{\includegraphics[width=\linewidth]{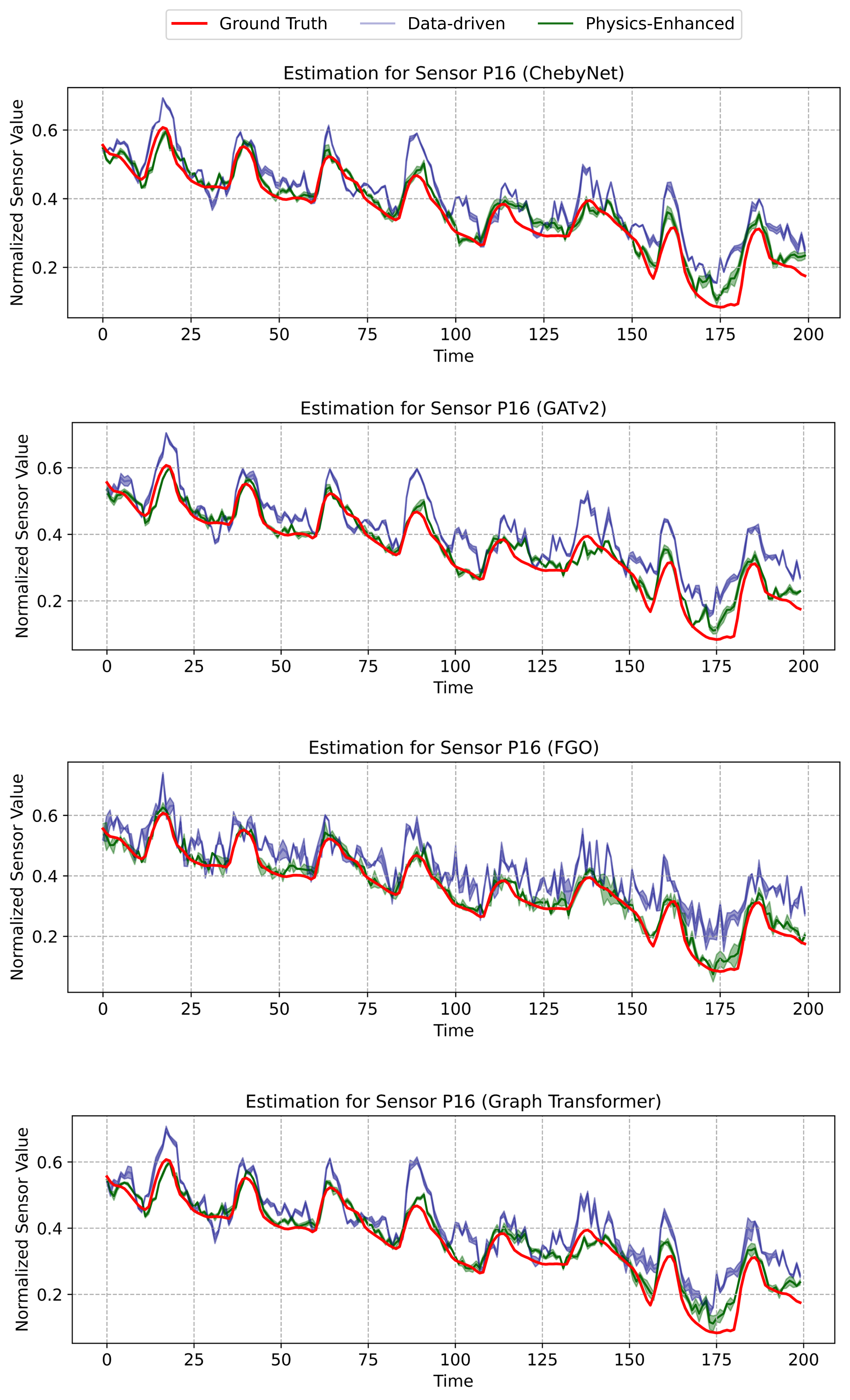}}
\caption{A comparison of virtual sensor estimation for pressure sensor (P16)  in a noisy scenario between the physics-enhanced and the purely data-driven GNNs.}
\label{noisy estimation}
\end{figure}

\begin{figure}
\centerline{\includegraphics[width=\linewidth]{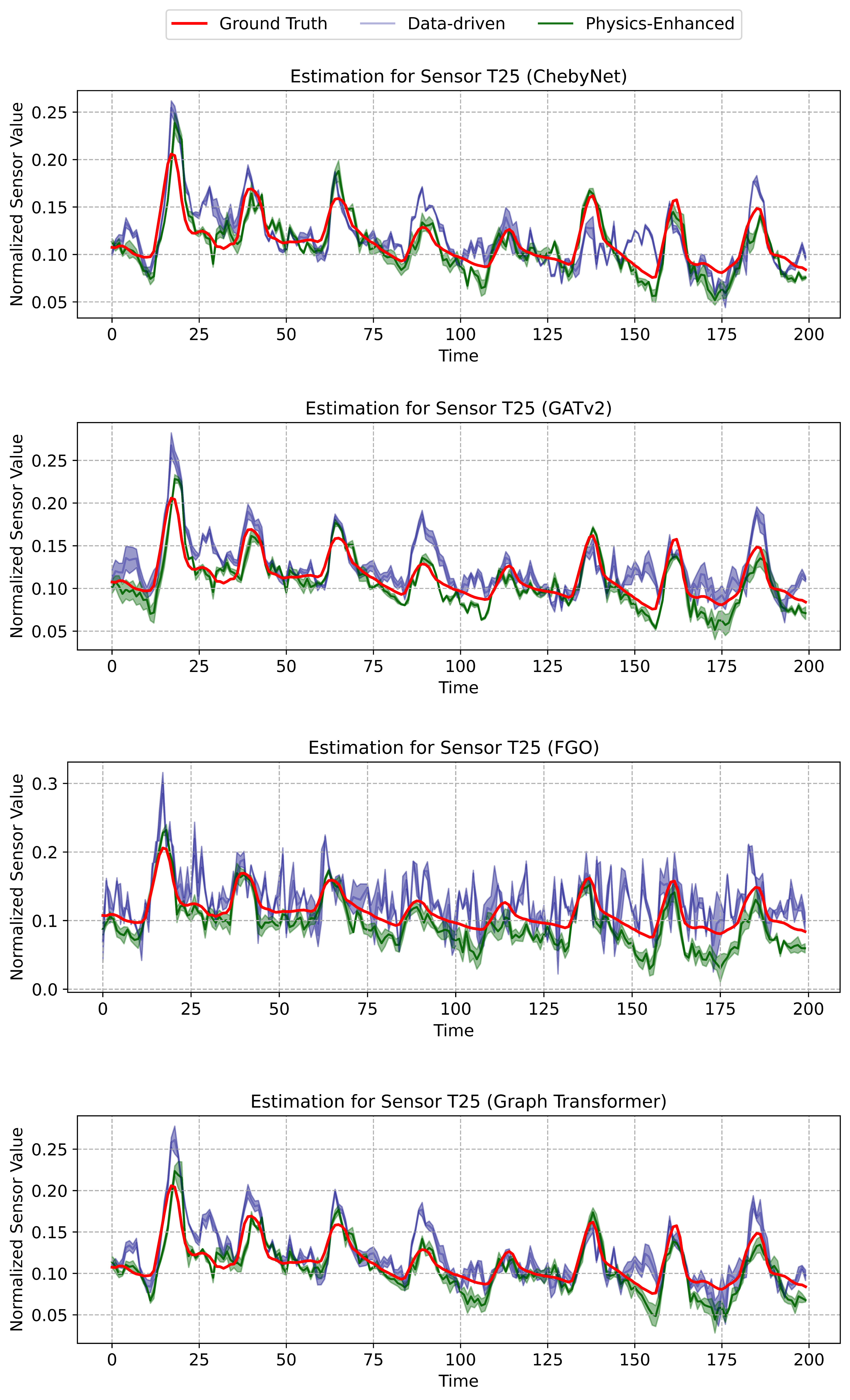}}
\caption{A comparison of virtual sensor estimation for temperature sensor (T25) in a noisy scenario between the physics-enhanced and the purely data-driven GNNs.}
\label{noisy estimation2}
\end{figure}

Table \ref{ideal} and Table \ref{noisy} present the performance of virtual sensor estimation using both the proposed physics-enhanced approach and the baseline approach (purely data-driven) under ideal (noise-free) and noisy conditions, respectively. Combining physics enhancement with any model yields superior results, with an average improvement of 51\% in terms of RMSE. Notably, the physics-enhanced Graph Transformer, in particular, demonstrates significant improvements, with relative reductions in RMSE and MAE   81\%, and 68\%, respectively, for ideal scenarios. For noisy cases, the physics-enhanced FGO shows the highest improvement with a 64\% relative reduction in RMSE. Additionally, accuracy increases for every physics-enhanced method, further illustrating the superiority of the proposed approach. Among the evaluated architectures, the four physics-enhanced GNNs (ChebyNet, GATv2, FGO, and Graph Transformer) consistently outperform other models. Therefore, these architectures offer an efficient solution for soft sensing, making them potent tools for data processing in district heating networks. In summary, the physics-enhanced approach, particularly when used with the Graph Transformer model, exhibits superior performance in noise-free scenarios. However, in noisy scenarios, GATv2 outperforms other models. Enhanced soft sensing performance is also evident when assessing individual virtual sensors. Figs.\ \ref{combined errorbar} and \ref{errorbar_noisy} illustrate the MAE for each sensor based on both purely data-driven and physics-enhanced graph-based models which have proven to be the best performing models for both noise-free and noisy cases, respectively. It is evident that the performance is significantly improved for almost all sensors in our experiments leading to overall model performance enhancement as illustrated in Fig \ref{fig: overalperformance}.

Figures \ref{noisy estimation} and \ref{noisy estimation2} illustrate the estimation performance of three graph-based models for particular pressure (P16) and temperature (T25) sensors in noisy scenarios, respectively. It is clear that the physics-enhanced framework outperforms solely data-driven approach across all GNN models, offering more accurate and reliable estimations. While purely data-driven GNN models utilize interactions among sensor measurements and historical data to learn mappings from mass flow rate values to temperature and pressure, they do not incorporate crucial information about  the physical process involved. In contrast, physics-enhanced models integrate insights on how water pressure and temperature of water change through the pipe, factoring in pipe characteristics and mass flow rate measurements. This integration of physical knowledge not only enriches the model's input but also has the potential to enable the understanding of evolving performance degradation (e.g. pipe leakage) within the district heating network—issues that are not explicitly modeled, resulting in significantly lower estimation errors than those seen in purely data-driven models. Furthermore, integrating physical knowledge helps the model to distinguish between normal operational variations and actual signs of degradation. In a district heating network, minor fluctuations in temperature and pressure are expected due to changes in demand and supply conditions. A purely data-driven model might misinterpret these fluctuations as anomalies, whereas a physics-enhanced model can recognize them as normal variations if they align with the expected physical behavior. Conversely, if the model detects a pattern of deviations that do not fit the physical expectations, it can flag this as a potential indicator of an underlying issue, such as a small leak that is gradually getting worse.

\subsection{Complexity Analysis}
In this subsection, we exclusively focus on the time complexity of the message passing and graph convolution operations within each model. For ChebyNet, GATv2, and Graph Transformer, computing the representation of each node $v_i$ involves considering its neighbors, with the sum of neighbors across all nodes precisely equaling the number of edges. Therefore, the general time complexity for these approaches is $\mathcal{O}(|\mathcal{E}|)$. In the case of the physics-enhanced model, augmentation of physics-based nodes ($|\mathcal{V}_{\text{aug}}|$) increases the number of augmented edges ($|\mathcal{E}_{\text{aug}}|$), resulting in linear growth in time complexity expressed as $\mathcal{O}(|\mathcal{E}| + |\mathcal{E}_{\text{aug}}|)$. In the extreme scenario where the adjacency matrix is fully connected (not applicable in our case due to distance thresholding in equation (\ref{graphconstruction}))), the complexity exhibits quadratic growth with respect to the number of nodes, denoted as $\mathcal{O}((|\mathcal{V}| + |\mathcal{V}_{\text{aug}}|)^2)$. For FGO, the computational complexity follows a log-linear relationship with respect to the number of nodes $(|\mathcal{V}^h|)$ in the hypervariate graph structure (details are provided in Appendix), denoted as $\mathcal{O}(|\mathcal{V}^h| \log{|\mathcal{V}^h|})$, whereas for physics-enhanced FGO model, the complexity is expressed as $\mathcal{O}((|\mathcal{V}^h| + |\mathcal{V}^h_{\text{aug}}|) \log{(|\mathcal{V}^h| + |\mathcal{V}^h_{\text{aug}}|)})$.

\section{Conclusion}
\label{sec:conclusion}

This study proposes a novel physics-enhanced framework that combines information from the network's component parameters with mass flow rate sensors to improve soft sensing capabilities. We leverage factors such as pipe diameters, pipe roughness, and pipe lengths, in combination with mass flow rate sensors, to extract valuable insights related to temperature and pressure drops. To create a reliable physics-enhanced virtual sensor estimation model, we concatenate the estimated physical process with available sensor readings as input for neural networks. We evaluate the performance of the proposed method using a synthetic dataset that simulates a district heating network and several architectures, with a focus on GNNs. Our results demonstrate that the physics-enhanced approach outperforms purely data-driven models, which rely solely on sensor data as input.

Another important finding from our experiments is the superior performance of GNNs when combined with physics-based inputs. This advantage arises from their ability to leverage neighboring information, including mass flow rate sensor data and physical attributes, thereby improving the network's virtual sensor estimation capabilities. Our proposed framework also exhibits robustness against noise stemming from sensor quality and the inaccuracy of physical parameters, as confirmed by our studies. 

As a result, this study offers a promising direction for further work in transferring the soft sensing-enabled IIoT to other networks such as electricity, water, and gas. Future work may explore scenarios where certain physical parameters remain inaccessible (e.g.,\ heat transfer rate) and reformulate separate models to initially estimate these missing physical parameters. Subsequently, these estimated values could be seamlessly integrated into our proposed physics-enhanced model for soft sensing.

\section*{Acknowledgement}
This research was funded by the Swiss Federal Institute of Metrology (METAS).

\appendix

\subsection{Details on Baselines}
\subsubsection{Chebyshev Spectral Graph Convolutional Operator}
To address the substantial computational complexity and non-local nature of the filter \(\mathbf{g}_{\theta}\), Defferrard \emph{et al}.\ \cite{defferrard2016convolutional} introduced Chebyshev polynomials as an approximation technique for \(\mathbf{g}_{\theta}\) based on the diagonal matrix of eigenvalues. Specifically, 

\begin{equation}
\mathbf{g}_{\theta} = \sum_{k=0}^{K} \theta_k T_k(\tilde{\mathbf{\Lambda}}),
\end{equation}
where, the scaled eigenvalue matrix is denoted as $\tilde{\mathbf{\Lambda}} = 2\mathbf{\Lambda}/\lambda_{\text{max}} - \mathbf{I_N}$ (where $\lambda_{\text{max}}$ is the maximum eigenvalue), $\theta_k$ are the Chebyshev coefficients, and $T_k(\tilde{\mathbf{\Lambda}})$ represents the diagonal matrix with diagonal entries as the Chebyshev polynomials of order $k$ applied to the scaled eigenvalues. Since $T_k(\tilde{\mathbf{L}}) = \mathbf{U} T_k(\tilde{\mathbf{\Lambda}}) \mathbf{U}^\mathsf{T}$, the structure of the Chebyshev Spectral Graph Convolutional Network (ChebyNet) can be expressed as:
\begin{equation}
\mathbf{x} *_{\mathcal{G}} \mathbf{g}_{\theta} = \sum_{k=0}^{K} \theta_k T_k(\tilde{\mathbf{L}})\mathbf{x},
\end{equation}
where $\tilde{\mathbf{L}} = 2\mathbf{L}/\lambda_{\text{max}} - \mathbf{I_N}$.

\subsubsection{Fourier Graph Operator} For a multivariate time series window \(\mathbf{X}_t \in \mathbb{R}^{N \times T}\) at timestamp \(t\), where \(N\) is the number of sensors and \(T\) is window size, hypervariate graph \(\mathcal{G}_t^T = (\mathbf{X}_t^T, \mathbf{A}_t^T)\) can be constructed \cite{yi2024fouriergnn}. This graph is initialized as a fully-connected network with \(n = NT\) nodes. Fourier Graph Operator (FGO) can be defined as $\mathbf{S}_{\mathbf{A},\mathbf{W}} = \mathcal{FT}(\mathbf{\xi})$, where \(\mathcal{FT}\) denotes the Discrete Fourier Transform (DFT) and $\mathbf{\xi}$ is green kernel such that $\mathbf{\xi}[i, j] := \mathbf{A}(i,j) \circ \mathbf{W}$ and $ \xi[i, j] = \xi[i - j]$. Using the convolution theorem, the multiplication of \( \mathcal{FT}(\mathbf{X}) \) and \( \mathbf{S}_{\mathbf{A},\mathbf{W}} \) in Fourier space can be written as:

\begin{equation}
\begin{aligned}
\mathcal{FT}(\mathbf{X})  \odot \mathcal{FT}(\xi) &= \mathcal{FT}((\mathbf{X} \ast \xi)[i]) = \mathcal{FT}\left(\sum_{j=1}^{n} \mathbf{X}[j] \xi[i - j]\right) \\
 \\
&= \mathcal{FT}\left(\sum_{j=1}^{n} \mathbf{X}[j] \xi[i, j]\right), \quad \forall i \in [n],
\end{aligned}
\end{equation}
where \( (\mathbf{X} \ast \xi)[i] \) is the convolution of \( \mathbf{X} \) and \(\xi\). Given $\mathbf{\xi}[i, j] := \mathbf{A}(i,j) \circ \mathbf{W}$, this yields:

\begin{equation}
\sum_{j=1}^{n} \mathbf{X}[j] \xi[i, j] = \sum_{j=1}^{n} \mathbf{A}(i,j) \mathbf{X}[j] \mathbf{W} = \mathbf{AXW},
\end{equation}

leading to the convolution equation:

\begin{equation}
\mathcal{FT}(\mathbf{X}) \odot \mathbf{S}_{\mathbf{A},\mathbf{W}} = \mathcal{FT}(\mathbf{AXW}).
\end{equation}

\subsubsection{Attention Mechanism for Graph Neural Networks}
Masked self-attentional layers for prediction have gained popularity for their ability to achieve state-of-the-art results \cite{vaswani2017attention}. 
Consequently, several models aim to extend the attention operator to graphs, where attention-based operators assign different weights to neighbors leading to improved results.

\subsubsection{Graph Attention Networks (GATs)} The core operation within GAT \cite{veličković2018graph} involves gathering features from neighboring nodes using an attention mechanism in the following way:

\begin{equation}
   \mathbf{h}^{\prime}_i = \sigma \left( \sum_{j \in \mathcal{N}(i)} \alpha_{ij} \mathbf{W} \mathbf{h}_j \right),
\end{equation}
where $\mathbf{W}$ denotes a weight matrix, $\mathbf{h}_j$ represents the feature vector of node $j$, $\alpha_{ij}$ represents the attention coefficient between node $i$ and node $j$, $\sigma$ is the non-linear activation function, and $\mathcal{N}(i) = \{j \in \mathcal{V} \, | \, e_{ij} \neq 0\}
$. The computation of $\alpha_{ij}$ is determined as follows:
\begin{equation}
    \alpha_{ij} = \text{softmax}_j \left( e(\mathbf{h}_i, \mathbf{h}_j)  \right) = \frac{\exp(e(\mathbf{h}_i, \mathbf{h}_j))}{\sum_{k \in \mathcal{N}(i)} \exp(e(\mathbf{h}_i, \mathbf{h}_k))}
    ,
\end{equation}
here, $e(\mathbf{h}_i, \mathbf{h}_j)$ corresponds to the unnormalized attention score between node $i$ and node $j$:
\begin{equation}
    e(\mathbf{h}_i, \mathbf{h}_j) = \text{LeakyReLU}\big(\mathbf{a}^\mathsf{T} \big[\mathbf{W}\mathbf{h}_i || \mathbf{W}\mathbf{h}_j \big]\big),
\end{equation}
where $||$ denotes the concatenation operator and $\mathbf{a}$ is a learnable parameter.

While the introduction of GAT marked an advancement in GNNs, it has been noted that the initial attention mechanism used in GAT, commonly referred to as ``static" attention, possesses certain limitations when it comes to effectively capturing complex information. This static attention maintains a consistent ranking of attention scores regardless of the characteristics of the query node. While this form of attention is effective in specific scenarios, it limits GATs in effectively addressing intricate graph-related challenges. To cope with this limitation, a more expressive variant named GATv2 has been introduced \cite{brody2022how}. GATv2 integrates a dynamic attention mechanism that allows nodes to adapt their attention dynamically based on their features and interactions with neighboring nodes by:
\begin{equation}
    e(\mathbf{h}_i, \mathbf{h}_j) = \mathbf{a}^\mathsf{T}\text{LeakyReLU}\big(\mathbf{W} \big[\mathbf{h}_i || \mathbf{h}_j \big]\big).
\end{equation}
Using the computed attention coefficients, a multi-head attention is formulated as:
\begin{equation}
\mathbf{h}'_i =\bigg\Vert_{k=1}^K \sigma \left( \sum_{j \in \mathcal{N}(i)} \alpha^{k}_{ij} {\mathbf{W}}^{k} \mathbf{h}_j \right),
\end{equation}
where $K$ denotes the number of multi-head attention.

\subsubsection{Graph Transformer} The Graph Transformer \cite{ijcai2021p214} is another variant of GNNs that leverages attention mechanisms during message passing. Its formulation is expressed as follows:

\begin{equation}
\mathbf{h}^{\prime}_i = \sigma \left(\mathbf{W}_1 \mathbf{h}_i +
\sum_{j \in \mathcal{N}(i)} \alpha_{ij} \mathbf{W}_2 \mathbf{h}_{j}\right).    
\end{equation}

Some differences can be observed compared to GATs. Firstly, the layer adds the transformed features of the central node to the output, which is denoted as 
$\mathbf{h}^{\prime}_i$. Furthermore, it transforms the source feature $\mathbf{h}_i$ and distant feature $\mathbf{h}_j$ into the query and key vectors, respectively, with different trainable parameters $\mathbf{W}_3$ and $\mathbf{W}_4$. Hence, the attention coefficient is calculated as follows:

\begin{equation}
\alpha_{ij} = \text{softmax}_j \left(
\frac{(\overbrace{\mathbf{W}_3\mathbf{h}_i} ^\textrm{query})^{\mathsf{T}} (\overbrace{\mathbf{W}_4\mathbf{h}_j} ^\textrm{key})}
{\sqrt{d}} \right),
\end{equation}
where $d$ represents the hidden size of each attention head.

\section*{Acknowledgement}
This research was funded by the Swiss Federal Institute of Metrology (METAS).

\bibliographystyle{IEEEtran}
\bibliography{IEEEabrv,main}

\end{document}